\pdfoutput=1

\documentclass[11pt]{article}

\usepackage{acl}

\usepackage{times}
\usepackage{latexsym}
\usepackage{graphics}
\usepackage{graphicx}
\usepackage{multirow}
\usepackage{placeins}
\usepackage[dvipsnames]{xcolor}
\usepackage{listings}
\usepackage{float}
\newfloat{listing}{tbp}{lol}
\floatname{listing}{Listing}
\usepackage[T1]{fontenc}

\usepackage[utf8]{inputenc}

\usepackage{microtype}

\usepackage{inconsolata}

\usepackage{url}

\usepackage{booktabs}

\newcommand{\metal}{CAT}

\title{Bidirectional Small-Granularity Search between Code and Text}

\author{Marco A. Valenzuela-Esc\'{a}rcega$^1$, Enrique Noriega-Atala$^2$,\\ \textbf{Gus Hahn-Powell$^2$}, \textbf{Clayton T. Morrison$^2$} \and \textbf{Mihai Surdeanu$^2$} \\
        $^1$Lex Machina\\
        $^2$The University of Arizona, Tucson, AZ, USA\\
        \texttt{mvalenzuela@lexmachina.com} \\ \texttt{\{enoriega,hahnpowell,claytonm,msurdeanu\}@arizona.edu}}

\begin{document}
\maketitle
\begin{abstract}

We introduce the novel task of bidirectional small-granularity search between code and text, where the queries are small snippets of text or code and the results are also small fragments of the opposite modality, i.e., code or text. This task establishes direct links between text in scientific publications and corresponding code segments, in support of better and faster understanding of scientific methods. We introduce a large dataset for the proposed task that includes a training partition with textual descriptions of code generated automatically using GPT-4, and three testing partitions, one in-domain and two out-of-domain (OOD) that contain manually-annotated data as well as material from other domains. 
We also propose a modular approach to address this task. Our approach shares an encoder across four different subtasks that learn start/end of answer spans in both directions. We show that our method achieves good results in-domain, and encouraging results OOD. This suggests that addressing this task with automatically-generated data is possible, but there is exciting future work to be done. 

\end{abstract}

\section{Introduction}

Recently, language models have expanded into modeling text and code jointly \cite{feng2020codebert,guo2021graphcodebert}. This has enabled several new applications such as code search and code generation from a textual specification \cite{10.1145/3368089.3417058,neelakantan2022text,wang2021syncobert}. 

While these are important tasks, we argue that one critical application is not properly addressed today: {\em bidirectional, small-granularity search} between code and text, where the queries are small snippets of text or code and the results are also small fragments of the opposite modality, i.e., code or text. We argue that this application addresses a critical need in the scientific domain where textual information that describes theoretical knowledge (e.g., publications, documentation, customer support documents) is loosely aligned with software repositories that provide practical implementations. 
For example, it is common that causal models in multiple domains (e.g., epidemiology, climate change) are described in publications at a theoretical level, while their implementation contains practical information such as values of various parameters and specific implementation details. 
By establishing direct links between the text in publications and specific code segments, subject matter experts (SMEs) can effortlessly correlate theoretical concepts with their practical implementation, reducing the time and effort required to understand how a model is structured and functions.





In this effort, we formally define such a task through both a novel dataset and several search approaches. In particular, the contributions of this paper are:

{\flushleft {\bf (1)}} We introduce a novel dataset for the proposed bidirectional small-granularity search task between text and code. The dataset includes four partitions. The first is a training partition that contains three hundred thousand pairs of aligned textual descriptions and code fragments. The code fragments come from two scientific domains: epidemiology and climate change. The textual descriptions were automatically generated using GPT-4. The second partition is for in-domain evaluation, i.e., it contains data from the same domains and textual descriptions generated in the same manner. Lastly, we include tow out-of-domain (OOD) evaluation partitions. The first contains software from the same domains associated with comments that were {\em manually annotated} as spans from the respective publications that describe the models. The second OOD partition contains data from a different domain (deep learning) together with manually-annotated comments. 
{\flushleft {\bf (2)}} We introduce a modular approach called \metal\ (from {\bf {\em C}}ode {\bf {\em A}}ligned with {\bf {\em T}}ext) to address this task. \metal\ is trained by maximizing the dot product 
between the encoding of the query and the encoding of the start/end answer tokens. \metal\ can use a variety of encoders such as CodeBERT or GraphCodeBERT, and a variety of training regimes that can incorporate in-batch negatives and label smoothing. Further, \metal\ supports a span indexing and search setting, in which the representations of spans in text/code sources are indexed using FAISS \citep{johnson2019billion} to support real-time search. 
{\flushleft {\bf (3)}} Lastly, we evaluate these approaches on the dataset described above to set initial results on this novel task. We analyze the behavior of our best approach to understand where it performs well and what its current limitations are to indicate potential directions for future work.


\section{Related Work}

\subsection*{Language Models}

Work on pre-training and fine-tuning large language models (LLMs) using a mixture of computer source code and natural language text have received considerable attention recently. Such efforts
include CodeBERT~\cite{feng2020codebert}, which is an encoder based on RoBERTa~\cite{liu2019roberta} that is pre-trained using code and text and fine-tuned on various NLP tasks. \citet{guo2021graphcodebert} expands on this idea with GraphCodeBERT, which leverages during pre-training the program data flow, i.e., a graph representation of the dependency relation between variables. 
To better understand  the advantage of incorporating source code in the LLM training, \citet{9609166} did a deep dive in the performance and properties of CodeBERT; their results support the  generalization of the model to unseen data and for new, downstream tasks.  

In addition to encoder architectures, there is also relevant research in decoder and sequence-to-sequence architectures for code generation. IntelliCode Compose~\cite{10.1145/3368089.3417058} is trained to predict sequences of code tokens of arbitrary types to generating up to entire lines of syntactically correct source code. PLBART~\cite{ahmad-etal-2021-unified}, also a sequence-to-sequence model, was trained in a multi-modal setting, which includes Python, Java and English language, for tasks such as summarization, translation, and generation of code. \citet{10.1145/3510003.3510062} proposed an approach that, instead of pre-training a network with source-code data, exclusively fine-tunes an existing natural language model such as RoBERTa with source code; they achieve results competitive with methods that pre-trained using source code.

Work diverging from the commonly used supervised learning methods to train code-to-code and code-to-text models includes CodeRL~\cite{NEURIPS2022_8636419d}, an architecture for program synthesis, which was developed using an actor-critic framework in an attempt to improve generalization to unseen data. \citet{10.5555/3495724.3497454} proposed an unsupervised approach to develop a \emph{transcompiler} among multiple programming languages.

\subsection*{Benchmarks}

Traditional, well-established NLP benchmarks typically focus on natural language tasks in either open-domain~\cite{wang-etal-2018-glue, 10.5555/3454287.3454581, gehrmann-etal-2021-gem} or domain-specific scenarios~\cite{tsatsaronis2015overview, 10.1145/3458754, emnlp-2019-bionlp}. Several new benchmarks have been proposed by the research community to evaluate the performance of code and multi-modal LLMs. CodeSearchNet~\cite{husain2020codesearchnet} consists of a dataset and evaluation of source-code retrieval. CodeXGlue~\cite{DBLP:journals/corr/abs-2102-04664} is a collection of 14 datasets, including CodeSEarchNet, with over 10 different programming languages to evaluate code-to-code and multi-modal downstream tasks specific to programming languages, e.g., code completion, refinement, Cloze test, and code search. HumanEval~\cite{chen2021codex} is an evaluation benchmark created to measure the functional correctness of programs synthesized from ``docstrings,'' i.e., short natural language descriptions of functions and methods. It relies on unit tests to automatically evaluate the correctness of synthesized source code for a small hand-curated set of programming problems.

\subsection*{Comparison with Existing Efforts}

Here we propose an architecture for multi-modal code retrieval and alignment, i.e., going to and from source code and natural language descriptions of source code. Our approach and relies on transformer encoders pre-trained on source code and natural language and fine-tuned for fine-grained, bidirectional search between code and text. To this end, we train neural networks with an objective function that can be used as a similarity metric for search. Similar to us, SynCoBERT~\cite{wang2021syncobert} encodes the semantics of source code into vector representation, and uses a contrastive loss function to learn representations suitable for code search and retrieval. UniXcode~\cite{guo-etal-2022-unixcoder} incorporates AST information for bidirectional code search and summarization. In \cite{neelakantan2022text} the authors leverage a contrastive objective to train a multi-modal dense retrieval model. 

Dense retrieval is often used as a prior step to question answering (QA) in a retriever-reader approach. \citet{lee2021dense} proposed a method to build efficient large scale dense representations for retriever-readers without the need for processing documents on-demand during inference by fine-tuning on reading comprehension datasets and using in-batch negative samples. 

Despite these similarities, 
our proposed architecture is different from these previous work by being specifically designed as a {\em bidirectional retriever-reader} (code-to-text and text-to-code) that can retrieve {\em small-granularity spans as answers}. It takes as query either a source code snippet or its natural language description, and returns a span containing its equivalent representation in the opposite modality.

\section{Approach}\label{sec:approach}

\metal\ is designed to create a bridge between scientific documents and code repositories.
It operates by embedding both text from documents and code from repositories into vector representations.
These embeddings are then stored in two distinct vector databases, facilitating the retrieval of related code or documentation based on user queries.

In this section, we describe the creation of the dataset that aligns text and code, as well as the \metal\ search architecture. 

\subsection{Dataset Creation: Code Repositories}


The current version of \metal{} specifically targets Python code repositories.
The processing of these repositories involves several tailored steps to prepare the Python code:
First, we scan each repository to identify Python files, typically with \verb|.py| extensions.
Files that are unit tests or ``dunder'' files (e.g., \verb|__init__.py|) are removed to focus on the primary codebase.
Python files that are too short, potentially indicating a lack of substantive content, are also excluded.
This criterion helps in focusing on files likely to contain significant code patterns.
We employ Tree-sitter,\footnote{\url{https://tree-sitter.github.io/}} a parser generator tool and an incremental parsing library, for removing comments and docstrings.
Further, some Python files are too long for direct processing with a transformer network. 
We handle these files using Tree-sitter for semantic chunking, which is performed at the level of class or function definitions.
This approach ensures that each code segment is a self-contained unit, potentially representing a complete thought or functionality, thereby maintaining the integrity and context of the code during the embedding process.

\subsection{Aligning Code and Text Data}

\begin{listing}[htb]
\centering
\begin{lstlisting}[frame=topline,
escapeinside=||,
breaklines=true,
framesep=2mm,
basicstyle=\scriptsize\ttfamily]
We then generate all possible combinations of the states in the Markov chain [1-2]. Two variables are initialized to zero, which will be used to accumulate the transition and trend values [3-4].    
\end{lstlisting}

\begin{lstlisting}[frame=lines,
escapeinside=||,
framesep=2mm,
basicstyle=\scriptsize\ttfamily,
language=Python]
1   combs = pd.Series(product(
2      arr_markov.index, (arr_markov.columns)))
3   value_trans = 0
4   value_trend = 0
\end{lstlisting}
\vspace{-2mm}
\caption{Example of a description generated using GPT-4 (top) aligned to text (bottom). 
}
\label{listing:alignment-example}
\end{listing}

The task at hand is to link small segments of Python code to corresponding explanatory text.
This level of granularity and specificity in mapping is not commonly found in existing datasets.
While there are numerous code-related datasets available~\cite[inter alia]{husain2020codesearchnet,DBLP:journals/corr/abs-2102-04664,pile,chen2021codex}, they typically operate at coarser granularity for applications such as code search, summarization, or classification.
That is, these datasets may include code along with comments or documentation, but they do not provide the kind of detailed, sentence-to-line mapping that \metal{} requires.

To solve this granularity issue at scale, we employ GPT-4~\cite{brown2020language, openai2023gpt4} to produce our training and in-domain evaluation dataset partitions. Formally, we use GPT-4 to generate explanations for segments of code, accompanied by a detailed mapping at the sentence-to-line level.
This mapping directly associates specific lines or parts of the code chunk with their respective explanatory text.
Generating synthetic data offers notable benefits, such as the ability to exercise control over the quality and specificity of the training data.
This approach enables the generation of varied examples, which may be underrepresented in existing datasets, thereby enhancing the robustness and generalizability of the model. Listing \ref{listing:alignment-example} shows an example of a generated explanation aligned to its corresponding source code.\footnote{More examples can be found in appendix~\ref{appendix:dataset}.}

We performed a qualitative analysis of the generated descriptions and found 85\% to be of good quality, i.e., the description narrates  the code's semantics accurately, and 15\% of acceptable quality, i.e., the descriptions are accurate but they stay too close to the actual code.\footnote{Details about this evaluation are described in appendix \ref{appendix:analysis}.}

To evaluate \metal's performance in realistic settings, we also generated two out-of-domain (OOD) dataset partitions, where textual information was manually annotated. 
The first dataset contains textual information that comes from existing publications and was manually aligned with the respective code, but uses a domain that is also included in training (climate change). The second OOD evaluation partition aligns text and code from a deep learning book~\cite{Surdeanu:24}. These two partitions allow one to evaluate increasingly more realistic scenarios, i.e., transitioning from automatically-generated text to manually-written text in the same domain (former partition), and moving from automatically-generated text to manually-written text in a different domain (latter partition).

\subsection{Dataset Statistics}\label{sec:dataset-stats}

We downloaded 627 Python GitHub repositories, 54 related to epidemiology modeling and 573 on climate change modeling. All repositories considered had licensing permitting reusing their source code. 

After downloading and aligning the resulting semantic chunks with LLM-generated comments, we created a dataset with 353,495 code-text pairs. Table \ref{tab:dataset} contains the breakdown of pairs in the dataset by domain and partition; 
Appendix \ref{sec:repositories} contains the keywords used to identify and download the repositories.

\begin{table}[htb]
\centering
\resizebox{\columnwidth}{!}{%
\begin{tabular}{ccccc}
    \toprule
                & Train  & Test    & Validation &Total \\
     \midrule
     Epidemiology        &  26,687 & 3,283    & 1550 & 31,520\\
     Climate    & 273,405 & 31,848   & 16,722 & 321,975\\
     \midrule
     Total      & 300,092 & 35,131   & 18,272 & 353,495\\
     \bottomrule
\end{tabular}%
}

\caption{Number of code-text pairs in the dataset. We present the number of pairs, per domain, per split and their marginalized totals.}\label{tab:dataset}
\end{table}


We also created two OOD evaluation datasets to facilitate the evaluation of \metal\ in realistic settings. Both datasets {\em manually} align scientific texts with corresponding code snippets. The first OOD is in the climate change domain, while the second comes from a deep learning textbook~\cite{Surdeanu:24}. Thus, the first dataset is to be used to investigate the transfer of the \metal\ model from automatic annotations to manual ones; the second dataset adds also a new domain not included in the training data. To detail, for the climate change domain, we annotated 94 text-code pairs from publicly available Jupyter notebooks related to weather and glacier modeling; for the deep learning domain, we annotated 147 pairs from code examples in textbook and their corresponding textual descriptions.

Appendices \ref{appendix:dataset} and \ref{sec:ood-annotations} include examples of code-text pairs from the three datasets.

\subsection{\metal\ Architecture}

\begin{figure}[t]
    \begin{center}
    \resizebox{1.1\columnwidth}{!}
    {\includegraphics{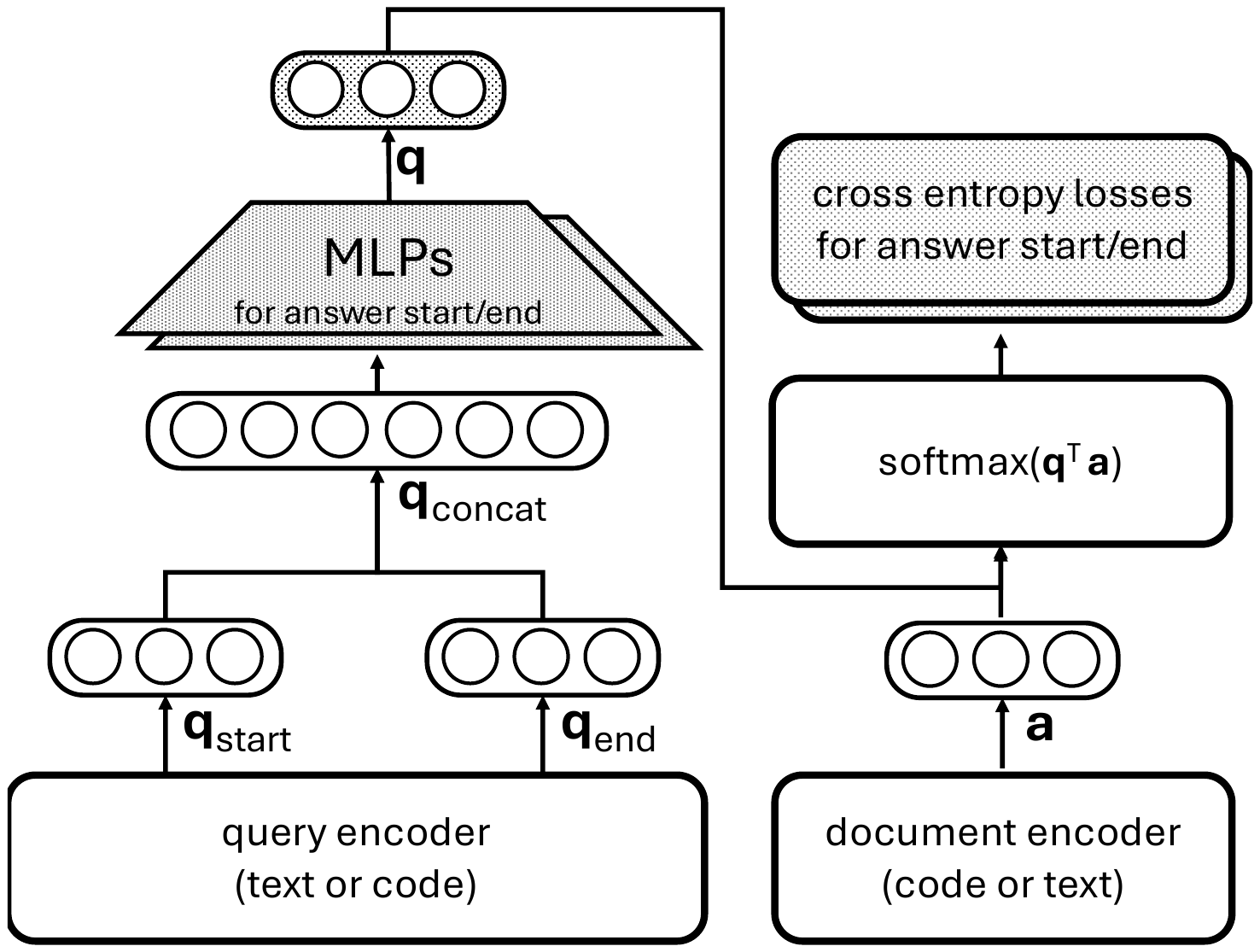}}
    \end{center}
    \vspace{-2mm}
    \caption{The \metal\ architecture. The components with a white background are shared between the two subtasks of predicting answer start/end. The components with a shaded background are unique to each subtask, i.e., there are distinct MLPs and losses for the two different subtasks. Further, the same architecture is applied in each direction, i.e., searching from text to code and vice versa.}
    \label{fig:arch}
    \vspace{-5mm}
\end{figure}

We propose an {\em encoder-based} approach for the CAT architecture. Encoders present key advantages over decoder-based LLMs for this task. First, encoding enables the use of dense information retrieval indices to rapidly find candidate passages for alignment. Performing a retrieval step prior to alignment removes the problem of the bounded context window size present in LLMs. That is, it allows one to search for the correct, opposite-modality passage in more documents than those that potentially fit within the limits of an LLM’s input length. Second, using encoder models is faster and cheaper than using decoder-based LLMs due to the reduced memory and compute requirements. 

Figure~\ref{fig:arch} depicts the \metal\ architecture. 
\metal\ encodes both the query and the document using a single encoder that has been pre-trained using both source code and natural language, e.g., 
GraphCodeBERT~\cite{guo2021graphcodebert}.

The process for encoding queries is adapted from \citet{lee2021dense}.
When encoding a query, \metal\ uses the embeddings for the first and last tokens of the selected query span ($\textbf{q}_{start}$ and $\textbf{q}_{end}$ in the figure).
The embeddings for the first and last tokens are then concatenated ($\textbf{q}_{concat}$).
This concatenated vector is then passed through a multi layer perceptron (MLP) to project it into a vector that has the same size of the answer token embeddings. This MLP contains two linear layers with a GELU activation function in between.
\metal\ uses two MLPs: one two predict the answer start token and one for the answer end token. 
In both subtasks, the output is a single, unified embedding ($\textbf{q}$) that represents the entire query span.

To identify the answer start/end tokens, this query embedding is compared against all answer token embeddings ($\textbf{a}$) using the dot product $\textbf{q}^T \textbf{a}$. The resulting value is converted to a probability using a softmax function. The entire network is trained using a joint cross entropy loss for two binary subtasks (i.e., to classify the answer start/end tokens). 
Further, the same encoder is applied in both directions, i.e., code-to-text and text-to-code, but the linear projections are specific to each direction. Thus, the trained \metal\ parameters after training include a fine-tuned encoder, used for both modalities and four MLPs (one for answer start and one for answer end in each direction). 


The training procedure can be customized with two important features: in-batch negatives and label smoothing. 
The effectiveness of in-batch negatives for learning dense representations conducive to retrieval has been demonstrated in previous research~\cite{karpukhin2020dense, lee2021dense, neelakantan2022embeddings}. 
Specifically, \citet{lee2021dense} constructs a $B \times B$ cosine similarity matrix for each batch, where $B$ represents the batch size.
This matrix is formed by comparing each query embedding in the batch (rows) with its respective correct token embeddings (columns).
The matrix's diagonal elements represent positive examples, providing $B-1$ negative examples for each query.
Our method diverges from this approach.
Instead, we use {\em all} tokens from {\em other} sequences in the same batch as negative examples. 
This strategy significantly increases the number of negative examples in a memory-efficient manner.

A second feature of the training process is label smoothing \cite{lukasik2020does,szegedy2016rethinking}. Label smoothing replaces the original ground truth label distribution in the cross entropy loss function with a mixture of the original label distribution with the uniform distribution $\mathcal{U}[0, K]$, where $K$ is the number of labels. This modification assigns a small probability mass to all the labels, which has a regularization effect that should improve the generalization of the model.

A key feature of the proposed method's bi-encoder architecture~\cite{lee2021dense} is the decoupling of answer span tokens (be it code or text segments) representations and query representations.
Due to this decoupled architecture, the token embeddings can be precomputed offline and then stored in a vector database for fast retrieval. During the inference phase, the embeddings corresponding to all tokens from all documents are considered as candidates. In particular,
we retrieve the answer start/end tokens with the highest similarity with the corresponding projection of the query embedding, under the following additional constraints: (a) the answer start/end tokens must come from the same document; (b) the position of the answer start token must be smaller than the position of the answer end token; and (c) the answer is too long, i.e. its length exceeds a pre-configured maximum length.


\section{Experimental Results}\label{sec:eval-metrics}


We use three evaluation settings to analyze the performance of the CAT architecture:

    \textbf{No Retrieval:} We present the model with a query, and let CAT select the the the answer {\em only} from the document of opposite modality containing the correct answer span. 
    With this setting we investigate the capability of the neural model to select a span that corresponds to the correct answer in the opposite modality. We report SQuAD 2 exact match and F1 score measures \cite{rajpurkar-etal-2016-squad}. The exact match score gives credit only to answers with a perfect matching span, whereas the F1 score is a ``softer'' criterion which gives credit to partial matches too. Table \ref{tab:results} shows the results for this setting.
    
    \textbf{Retrieval:} We do an initial retrieval step to choose the top context passage from a MIPS index \cite{johnson2019billion} that contains passages constructed using a sliding window. In our experiments we used a window size of 512 sub-word tokens, which is the maximum input size allowed by the underlying transformer, and a stride size of 384 tokens. Since the results are selected from individual passages, we indexed overlapping passages to reduce the likelihood of missing an answer because it is split between two adjacent passages. Then, we present that context passage to the CAT classifier to select an answer span from it. This setting reflects a realistic scenario in which we do not know ahead of time which document contains the answer. 
    This setting relies on the same SQuAD 2 exact match and F1 evaluation measures.
    Table \ref{tab:retrieval-results} shows the results for the retrieval setting.
    
    \textbf{Weighted Retrieval:} This setting is identical with the retrieval setting above, but uses measures tailored for ranking. The motivation for this configuration is that the original SQuAD 2 measures inspect only the top answer extracted, and give no credit to better answers that are ranked lower. To address this limitation, we propose variants of exact match and F1 that are weighted by the inverse of the rank of the answer position (similar to the mean reciprocal rank formula~\cite{manning2009introduction}). For example, the weighted form of exact match is computed using the formula: $\min_i \frac{1}{i} \textrm{exact\_match}(\textrm{answer}_i)$, where $i$ iterates over all ranked answer positions. The weighted F1 score is computed with a similar formula. Table \ref{tab:weighted-retrieval-results} shows ranked retrieval results.
    

\begin{table*}[t]
    \centering
    \small
    \begin{tabular}{clccccc}
    \toprule
           &   &       & \multicolumn{2}{c}{Code-to-Text} & \multicolumn{2}{c}{Text-to-Code} \\
         Model & Domain & Total & Exact Match  & F1 & Exact Match  & F1 \\\midrule
         
        \multirow{3}{*}{GCB+Sub} & Test Set & 5978 & 81.97\% & 89.21\% & 64.07\% & 77.40\%\\
         & Deep Leaning & 147 & 19.72\% &  41.56\% & 17.68\% &  31.00\% \\
         & Climate Change & 94 & 17.02\% & 43.20\% & 19.15\% & 35.05\%\\\midrule
         \multirow{3}{*}{GCB} & Test Set & 5978 &80.62\% & 88.00\% & 62.47\% & 76.18\%\\
         & Deep Leaning & 147 & 16.32\% &  34.18\% & 20.40\% &  32.18\% \\
         & Climate Change & 94 & 14.89\% & 38.25\% & 20.21\% & 35.52\%\\\midrule
         \multirow{3}{*}{CB} & Test Set &5978 &66.41\% & 78.95\% & 44.08\% & 57.09\%\\
         & Deep Leaning & 147 & 5.44\% &  18.14\% & 6.12\% &  8.08\% \\
         & Climate Change & 94 & 7.44\% & 27.82\% & 17.02\% & 31.64\%\\         
         \bottomrule
    \end{tabular}
    \caption{Bidirectional search evaluation using SQuAD 2 evaluation measures, using the {\em no retrieval} setting, i.e., searching for spans in the correct file only.}
    \label{tab:results}
\end{table*}

         

\begin{table*}[t]
    \centering
    \small
    \begin{tabular}{clccccc}
    \toprule
          &    &       & \multicolumn{2}{c}{Code-to-Text} & \multicolumn{2}{c}{Text-to-Code} \\
         Model & Domain & Total & Exact Match  & F1 & Exact Match  & F1 \\\midrule
         
         \multirow{3}{*}{GCB+Sub} & Test Set & 5978 & 68.40\% & 77.78\% & 44.90\% & 59.06\%\\
        & Deep Leaning  & 147 & 12.24\% &  29.31\% & 8.84\% &  17.13\% \\ 
         & Climate Change & 94 & 9.57\% & 28.09\% & 11.70\% & 23.98\%\\\midrule
         \multirow{3}{*}{GCB} & Test Set & 5978 &65.05\% & 74.45\% & 33.20\% & 46.52\%\\
         & Deep Leaning & 147 & 6.12\% &  14.06\% & 8.16\% &  16.56\% \\
         & Climate Change & 94 & 7.44\% & 19.05\% & 7.44\% & 18.55\%\\\midrule
         \multirow{3}{*}{CB} & Test Set & 5978 & 33.28\% & 49.91\% & 20.32\% & 34.46\% \\
         & Deep Leaning & 147 & 5.44\% &  18.19\% & 2.72\% &  7.03\% \\
         & Climate Change & 94 & 5.31\% & 13.15\% & 4.25\% & 10.98\%\\    
         \bottomrule
    \end{tabular}
    \caption{Bidirectional search evaluation using SQuAD 2 evaluation measures, using the {\em retrieval} setting, i.e., searching for spans in the entire document collection.}
    \label{tab:retrieval-results}
\end{table*}

         

\begin{table*}[t]
    \centering
    \small
    \begin{tabular}{clccccc}
    \toprule
           &   &       & \multicolumn{2}{c}{Code-to-Text} & \multicolumn{2}{c}{Text-to-Code} \\
         Model & Domain & Total & Exact Match  & F1 & Exact Match  & F1 \\\midrule
         
         \multirow{3}{*}{GCB+Sub} & Test Set & 5978 & 73.50\% & 80.66\% & 49.30\% & 63.21\%\\
         & Deep Learning & 147 & 15.44\% &  33.32\% & 14.37\% &  23.47\%\\
         & Climate Change & 94 & 13.79\% & 34.05\% & 17.83\% & 29.29\%\\\midrule
         \multirow{3}{*}{GCB} & Test Set & 5978 &70.50\% & 77.70\% & 37.66\% & 51.42\%\\
         & Deep Leaning & 147 & 8.63\% &  19.04\% & 12.85\% &  21.44\% \\
         & Climate Change & 94 & 11.03\% & 25.14\% & 13.66\% & 22.66\%\\\midrule
         \multirow{3}{*}{CB} & Test Set & 5978 &42.45\% & 55.20\% & 27.35\% & 41.33\%\\
         & Deep Leaning & 147 & 9.09\% &  23.01\% & 6.49\% &  13.13\% \\
         & Climate Change & 94 & 6.20\% & 16.63\% & 10.36\% & 18.05\%\\    
         \bottomrule
    \end{tabular}
    \caption{Bidirectional search evaluation using the {\em weighted retrieval} setting. Here we report a ranking version of the SQuAD 2 evaluation measures (see Section \ref{sec:eval-metrics}). }
    \label{tab:weighted-retrieval-results}
\end{table*}

         

\subsection{Main Results}

We trained a CAT model\footnote{We used the HuggingFace Transformers for our experiments.} based on GraphCodeBERT \cite{guo2021graphcodebert} with the dataset described in section \ref{sec:dataset-stats}. During training, we used in-batch negative examples and label smoothing. A consequence of using in-batch negative examples is that for any given text or code segments in the batch, there is just one opposite modality segment, the rest of them induce negative pairs. This severely skews the amount of negative examples used during training. To mitigate this, we sub-sampled the negative pairs during training.\footnote{We used a simple heuristic to keep equal amounts of code and text segments that don't align with their opposite in the current batch. The sample size equals the minimum between the number of unaligned text and unaligned code segments.}
This model is denoted as \texttt{GCB + Sub}\footnote{The best configuration used a learning rate of $5\times10^{-5}$ with linear learning rate decay and 1k warm up steps; Optimized  Adam $\beta_1=0.9$, $\beta_2=0.999$; Batch size of 28 and trained for three epochs on an A100 GPU with 40 GB of VRAM.} in Tables~\ref{tab:results}, \ref{tab:retrieval-results}, and~\ref{tab:weighted-retrieval-results}. All of the reported results are the average over the corresponding data set.

To understand the contribution of the various components in \metal, we also evaluated two ablative configurations. 
The first ablative model was trained without sub-sampling in-batch negative  examples to understand its effect in the model's performance (\texttt{GCB} in the tables). The second ablative model was trained without sub-sampling and using CodeBERT~\cite{feng2020codebert} as the underlying encoder (\texttt{CB} in the tables).

We draw the following observations from the results shown in the three tables:

{\flushleft {\bf (1)}} The complete model (\texttt{GCB + Sub}) achieves high F1 scores in the in-domain test set: 89.21\% for code-to-text and 77.4\% in for text-to-code. We observed that this is an asymmetrical task, i.e., it is more challenging to search from text to code than vice-versa. This pattern consistently holds through almost all the results reported in this work.
We hypothesize that removing comments from source code during training but keeping them during testing and inference in OOD might be partially responsible for this asymmetry.

{\flushleft {\bf (2)}} The difference in performance between the {\em no retrieval} settings (Table~\ref{tab:results}) and the retrieval-based ones (Tables~\ref{tab:retrieval-results} and~\ref{tab:weighted-retrieval-results}) is not that large. This indicates that the contextualized embeddings learned by the encoder indeed capture the semantic alignment between text and code, and this alignment is accessible through cosine-similarity search. 

{\flushleft {\bf (3)}} The out-of-domain (OOD) performance of the complete model is substantially lower. Overall, the performance on the climate change OOD partition is higher than on the deep learning partition. This is expected, as climate change data (albeit with automatically-generated texts) is included in the training data. However, our error analysis (section \ref{sec:error-analysis}) indicates that these results are conservative, and real-world performance of \metal\ is likely to be much higher. 

\subsection{Ablation Experiments Discussion}
We trained two ablated models to understand the contribution of different elements of the CAT architecture to its performance (shown in same tables). Namely, we are interested in the effects of sub-sampling negative examples and the choice of the underlying transformer model. 

The first model ({\tt GCB}) removes negative sub-sampling from the full model. Its performance is lower than that of the full model. 
The performance differences are larger on the OOD datasets as well as on the retrieval settings. This suggests that sub sampling has a regularization effect with a small effect on in-domain data and a more apparent one when generalizing to data unseen during training.

The second model (\texttt{CB}) was trained without sub-sampling and using CodeBERT \cite{feng2020codebert} as the underlying encoder. The model's results are considerably lower in general, indicating that training with a stronger backbone encoder matters. 
\section{Qualitative Error Analysis}\label{sec:error-analysis}

To better understand the real-world behavior of \metal, we performed a qualitative error analysis of the OOD evaluation on the deep learning book partition. To this end, we manually analyzed 40 errors in the text-to-code setting, and 40 errors in the code-to-text setting. To isolate the behavior of the neural span classifier, we performed this analysis in the {\em no retrieval} setting, i.e., when the classifier was pointed to the correct doc in which it had to identify the correct answer span. 

\begin{table}
    \centering
    \small
    \resizebox{0.7\columnwidth}{!}{%
    \begin{tabular}{ccc}
    \toprule
    & \multicolumn{2}{c}{Percentage} \\
    Error Type & Text-to-Code & Code-to-Text\\
    \midrule
    Comment & 42.5\% & - \\
    Correct & 22.5\% & 32.5\% \\
    Topical & 15.0\% & 15.0\% \\
    Overlap & 7.5\% & 10.0\% \\
    Other & 12.5\%  & 42.5\%\\
    \bottomrule
    \end{tabular}
    }
    \caption{Error distributions for text-to-code and code-to-text classification (i.e., {\em no retrieval}) on the OOD partition.}
    \label{tab:erroranalysis-texttocode}
\end{table}



The results of this analysis are summarized in Table~\ref{tab:erroranalysis-texttocode} for the two directions (text-to-code and code-to-text). We identified the following errors:

{\flushleft {\bf Comment:}} This error applies solely for text-to-code classification. In this situation the model extracted an answer span that is inside a comment present in the code rather than extracting an actual code span. 
For example, in one instance, the system extracted the comment {\em ``\# update the weights''} instead of the associated code {\tt w = w - lr * deriv\_cost}.

{\flushleft {\bf Correct:}} The answer span differs from the gold span in the original annotation but is actually correct according to the human evaluator. 
This happened due to redundancy in the answer document, or due to the fact that the system prefers longer spans that are more coherent and do include the correct span. 
For example, instead of the annotated correct answer {\em ``if the prediction is incorrect, then we need to adjust w and b''} the system extracted the complete sentence, which, arguably, is a more coherent output: {\em ``However, if the prediction is incorrect, then we need to adjust w and b, as described in Algorithm 2.''}

{\flushleft {\bf Topical:}} The answer span is incorrect but is topically related to the query and spatially close to the correct answer. For example, in response to the query {\em ``If the prediction is correct, then no update is needed, and we can move on to the next training example.''} the system extracted another stop condition in the same loop {\tt if n\_errors == 0: break}.

{\flushleft {\bf Overlap:}} The answer span is incorrect but it overlaps with the correct answer span. For example, in one instance the system extracted the line of code {\tt train\_df.insert(1, 'class', classes)}, which is only one of the three lines annotated as the correct code span.

{\flushleft {\bf Other:}} Other types of error that are hard to classify. 

The {\em Comment} error is an artifact of our training process, in which we removed comments from code when training the text-to-code model, to encourage the model to focus on code. However, the OOD evaluation did not change the code in any way to keep the evaluation realistic. Due to this, at evaluation time the model was exposed to data it has not seen in training.

Interestingly, the {\em Correct} situation was common: 22.5\% of errors for text-to-code and 32.5\% of errors for code-to-text. The larger percentage for code-to-text is due to the increased verbosity of text, which allowed for more redundancy. However, the same verbosity had the undesired side effect of increasing the number of false positives (shown under {\em Other}). 

The {\em Overlap} errors indicate that, even when the system is incorrect, its answers remain in close proximity to the correct answer. This, combined with the high percentage of {\em Correct} situations suggests that the real-world performance of \metal, e.g., when deployed within a search user interface, would be considerably higher than what we reported in our formal evaluation.

\section{Conclusion}

We introduced a novel task that addresses bidirectional small-granularity search between code and text, where the queries are small snippets of text or code and the results are also small fragments of the opposite modality, i.e., code or text. We accompany this effort with a large dataset that: (a) was created automatically from real Python repositories and textual descriptions generated using GPT-4, but (b) contains both in-domain and out-of-domain evaluation partitions that were manually annotated in multiple scientific domain. We also proposed a modular approach that addresses the task using a multi-task learning architecture that shares a single encoder for efficiency. Our initial results suggest that this task can be addressed with automatically-generated training data, but the task is not solved.


\section*{Limitations}

Although our approach is language independent (both with respect to natural and programming languages), we have evaluated our method only on English texts and Python software. It is possible that results on other combinations are lower, especially considering that most programming languages use English-based keywords that simplify alignment with English texts. 

\section*{Ethical Considerations}

We were careful to use only software repositories with permissive licenses, excluding viral ones such as GPL and even LGPL (see Table~1). Similarly, we included only open-access text documents. As such, the dataset created can be publicly released; we plan to do it immediately, if the paper is accepted.

Further, it is important to note that our method is extractive not generative, i.e., we use only encoders rather than decoder-based methods. Thus, our approach cannot be used for ``copyright washing'' of existing code.

Lastly, we currently focus only on documents written in English, which we acknowledge as exclusionary. We plan to incorporate other languages in future work. 

\bibliography{anthology,custom}

\appendix
\section{Source Repositories Retrieval}\label{sec:repositories}

We used GitHub's API\footnote{\url{https://docs.github.com/en/rest?apiVersion=2022-11-28}} to search domain-related repositories and download the archived versions. The repositories used to create the dataset contain an open-source license permissive enough to redistribute derivative works. Table~\ref{tab:licenses} shows the distribution of licenses included in these repositories.

\begin{table}[htb]
    \centering
    \resizebox{\columnwidth}{!}{%
    \begin{tabular}{lr}
\toprule
 License Name & \# Repositories \\
\midrule
MIT License & 472 \\
Apache License 2.0 & 97 \\
BSD 3-Clause "New" or "Revised" License & 38 \\
The Unlicense & 9 \\
Creative Commons Zero v1.0 Universal & 4 \\
BSD 2-Clause "Simplified" License & 3 \\
MIT No Attribution & 2 \\
Boost Software License 1.0 & 1 \\
Do What The F*ck You Want To Public License & 1 \\
\bottomrule
\end{tabular}
}
    \caption{Software licenses of the dataset repositories.}
    \label{tab:licenses}
\end{table}

For climate change, we used the following key phrases: \texttt{earth science, climate change, erosion, water erosion, wave erosion, wind erosion, glacial erosion, erosion by gravity, deposition, water deposition, wave deposition, wind deposition, glacial deposition, deposition by gravity, loss of soil, air pollution, water pollution, soil pollution, land pollution, weather, weather forecast, atmosphere, atmospheric layer, ocean current, surface water, groundwater, carbon cycle, nitrogen cycle}.

For epidemiology modeling, we used the following key phrases: \texttt{epidemiology, contact tracing, compartmental model, JHUAPL-Bucky}.

\section{GPT-4-Augmented Dataset}\label{appendix:dataset}
Listing \ref{listing:source} shows an example of a Python source code file\footnote{This source code file was originally pulled from \url{https://github.com/Unidata/MetPy/blob/main/examples/calculations/Advection.py} and striped of comments.} from the dataset described in section \ref{sec:approach}. Listing \ref{listing:text} contains a verbose description of the aforementioned Python code in English. We used GPT-4 to generate text descriptions of source code files in our dataset using the prompt template shown in listing \ref{listing:prompt}.

\begin{listing*}[!ht]
\begin{lstlisting}[escapeinside=||,
frame=lines,
framesep=2mm,
basicstyle=\footnotesize\ttfamily,
numbers=left,numberstyle=\tiny\color{gray},
language=Python]
import matplotlib.pyplot as plt

import metpy.calc as mpcalc
from metpy.cbook import example_data

ds = example_data()

tadv = mpcalc.advection(ds.temperature, ds.uwind, ds.vwind)

print(tadv.data.units)

fig, ax = plt.subplots(figsize=(5, 5))

|\colorbox{yellow}{cs = ax.contour(ds.lon, ds.lat, ds.temperature, range(4, 26, 2), colors=\textquotesingle{}tab:red\textquotesingle{},}|
|\colorbox{yellow}{                linestyles=\textquotesingle{}dashed\textquotesingle{}, linewidths=3)}|
plt.clabel(cs, fmt='%d', fontsize=16)

cf = ax.contourf(ds.lon, ds.lat, tadv.metpy.convert_units('kelvin/hour') * 3, range(-6, 7, 1),
                 cmap=plt.cm.bwr, alpha=0.75)
plt.colorbar(cf, pad=0, aspect=50)
ax.barbs(ds.lon.values[::2], ds.lat.values[::2],
         ds.uwind[::2, ::2], ds.vwind[::2, ::2],
         color='black', length=6, alpha=0.5)
ax.set(xlim=(260, 270), ylim=(30, 40))
ax.set_title('Temperature Advection Calculation')

plt.show()
\end{lstlisting}
\caption{Example source code file used to generate descriptions for the dataset. The code highlighted in lines 14 and 15 correspond to the description highlighted in listing \ref{listing:text}.}
\label{listing:source}
\end{listing*}

\begin{listing*}[!ht]
\begin{lstlisting}[escapeinside=||,
breaklines=true,
frame=lines,
framesep=2mm,
basicstyle=\footnotesize\ttfamily,
numbers=left,numberstyle=\tiny\color{gray}]
We begin by importing the necessary libraries for our task. We import matplotlib.pyplot, which is a plotting library used for creating static, animated, and interactive visualizations in Python. We also import specific functions from the metpy library, which is a collection of tools in Python for reading, visualizing, and performing calculations with weather data.

Next, we load an example dataset using the example_data function from the metpy.cbook module. This dataset contains various meteorological parameters such as temperature, wind speed, and wind direction.

We then calculate the temperature advection using the advection function from the metpy.calc module. Temperature advection is a meteorological term that refers to the rate of change of temperature due to the horizontal movement of air. The advection function takes three arguments: the temperature, and the u and v components of the wind. The result of this calculation is stored in the variable tadv.

We print the units of the calculated temperature advection data. This is important as it helps us understand the scale and dimension of the data we are working with.

We then proceed to visualize our data. We create a new figure and axes with a specified size using the subplots function from matplotlib.pyplot. |\colorbox{yellow}{We plot contours of the temperature data on these axes using the contour function.}| The contour lines are labeled with their respective values using the clabel function.

We also create a filled contour plot of the temperature advection data, which has been converted to units of kelvin per hour and multiplied by 3. A colorbar is added to the plot for reference. We then plot wind barbs on the same axes to represent the wind speed and direction. The wind barbs are plotted at every other point in the dataset to avoid overcrowding the plot.

We set the limits of the x and y axes to specific values and add a title to the plot. Finally, we display the plot using the show function from matplotlib.pyplot. This plot provides a visual representation of the temperature advection calculation, allowing us to better understand the spatial distribution of this meteorological phenomenon.
\end{lstlisting}
\caption{Example of LLM-generated text description of the source code from listing \ref{listing:source}. The phrase highlighted in line 9 describes the instructions highlighted in listing \ref{listing:source}.}
\label{listing:text}
\end{listing*}

\begin{listing*}[!ht]
\begin{lstlisting}[escapeinside=||,
breaklines=true,
frame=lines,
framesep=2mm,
basicstyle=\scriptsize\ttfamily,
numbers=left,numberstyle=\tiny\color{gray}]
In this annotation task, you will be asked to describe the high-level semantics
of a provided code in the form of a story. The story should be self-contained,
written in academic terms, and thoroughly cover all aspects of the code semantics.
You will annotate each sentence with the corresponding line numbers of the code,
following the instructions described below.

You will annotate the code provided to you, which includes a series of functions,
variables, and libraries. Your annotation should describe the purpose and intuition
behind each of these elements, without mentioning their names.

When annotating your story, each sentence should describe a section of the code,
and include the correct line numbers surrounded by square brackets after the complete
corresponding phrase. Here are some examples of valid annotations:

    [28] refers to line 28
    [12-19] refers to lines 12 to 19

Note that commas are not allowed inside an annotation.
For example, [29-32, 35-36] is invalid.

When referring to the actions performed by the code, you should use the pronoun "we."
For example, "We initialize a dictionary" instead of "The code initializes a dictionary."

Here is an example of a sentence with the corresponding annotation:

    Sentence:
    We set the initial values for the air temperature and the relative humidity,
    and then we calculate the heat index.

    Annotated sentence:
    We set the initial values for the air temperature [5] and the relative humidity [6],
    and then we calculate the heat index [8].

    Annotations:
    [5] refers to the line of code where the air temperature is specified.
    [6] refers to the line of code where the relative humidity is specified.
    [8] refers to the line of code where the heat index is calculated.

    Sentence:
    We define a constant that is the ratio of a circle's circumference to its diameter.

    Annotated sentence:
    We define a constant that is the ratio of a circle's circumference to its diameter [13].

    Annotations:
    [13] refers to the line of code where the constant is defined.

{%- if description -%}
### REPOSITORY DESCRIPTION ###

{{ description }}
{%- endif -%}

### CODE ###

{{ code }}

### STORY ###

\end{lstlisting}
\caption[Prompt template for GPT-4 code description generation.]{Prompt template used to generate descriptions of a source code file using GPT-4. We used LangChain\footnote{\url{https://python.langchain.com/docs/modules/model_io/prompts/prompt_templates/}} to generate a prompt for each source code file in the dataset. The code is interpolated into the prompt in line 56. If the repository contains a description, we interpolate it too (lines 48--52), to prime the LLM with context information to generate more accurate descriptions.}
\label{listing:prompt}
\end{listing*}

\begin{listing*}[!ht]
\begin{lstlisting}[escapeinside=||,
frame=lines,
framesep=2mm,
basicstyle=\footnotesize\ttfamily,
numbers=left,numberstyle=\tiny\color{gray},
language=Python]
lr = 1e-1
n_epochs = 10

indices = np.arange(n_examples)
for epoch in range(10):
    # randomize the order in which training examples are seen in this epoch
    np.random.shuffle(indices)
    # traverse the training data
    for i in tqdm(indices, desc=f'epoch {epoch+1}'):
        x = X_train[i]
        y = y_train[i]
        # calculate the derivative of the cost function for this batch
        |\colorbox{yellow}{deriv\_cost = (sigmoid(x @ w) - y) * x}|
        # update the weights
        w = w - lr * deriv_cost

\end{lstlisting}
\caption{Source code example from the deep learning dataset. The code highlighted in line 13 implements the description from listing \ref{listing:text-dl}.}
\label{listing:source-dl}
\end{listing*}

\section{Out-of-domain Annotations for Evaluation}
\label{sec:ood-annotations}
Listings \ref{listing:source-dl} and \ref{listing:text-dl} show an example of a code-description pair used to evaluate our method in the domain of \emph{deep learning}. Similarly, listings \ref{listing:source-climate} and \ref{listing:text-climate} show an example pair used to evaluate our method in the domain of \emph{climate change}.

\begin{listing*}[!ht]
\begin{lstlisting}[escapeinside=||,
breaklines=true,
frame=lines,
framesep=2mm,
basicstyle=\footnotesize\ttfamily,
numbers=left,numberstyle=\tiny\color{gray}]
This function can be easily implemented in NumPy as follows: 

However, this naive implementation may produce the following warning during training: 

The term overflow indicates that the result of evaluating exp(-x) is a number so large that it can't be represented by a float (specifically, we're using float64 numbers).
We will avoid this issue by not calling exp with values that will overflow.
NumPy provides the function finfo that can be consulted to find the limits of floating point numbers: 

The log of the largest floating point number is the largest number for which exp() will not overflow, so we will use it as a threshold to filter out problematic values: 

We now have everything we need to implement Algorithm 4.
The steps to follow for each example are: (1) use the model to make a prediction, (2) |\colorbox{yellow}{calculate the gradient of the loss function with respect to the model parameters}|, and (3) update the model parameters using the gradient.
The size of the update is controlled by the learning rate. 

Once the model has been trained, we evaluate it on the test dataset using our binary_classification_report function from the previous section.
\end{lstlisting}
\caption{Text snippet example from the deep learning dataset. The phrase highlighted in line 12 describes the instructions highlighted in listing \ref{listing:source-dl}.}
\label{listing:text-dl}
\end{listing*}

\begin{listing*}[!ht]
\begin{lstlisting}[escapeinside=||,
frame=lines,
framesep=2mm,
basicstyle=\footnotesize\ttfamily,
numbers=left,numberstyle=\tiny\color{gray},
language=Python]
bed.plot()

bed

ELA = 3000 # equilibrium line altitude in meters above sea level
altgrad = 4 # altitude gradient in mm/m
mb_model = MassBalance(ELA, gradient=altgrad)

mb_model

|\colorbox{yellow}{glacier = Glacier(bed=bed, mass\_balance=mb\_model)}|

glacier

glacier.plot_mass_balance()

runtime = 1
glacier.progress_to_year(runtime)
\end{lstlisting}
\caption{Example code snippet from the climate change dataset. The code highlighted in line 11 implements the description from listing \ref{listing:text-climate}.}
\label{listing:source-climate}
\end{listing*}

\begin{listing*}[!ht]
\begin{lstlisting}[escapeinside=||,
breaklines=true,
frame=lines,
framesep=2mm,
basicstyle=\footnotesize\ttfamily,
numbers=left,numberstyle=\tiny\color{gray}]
 linear mass balance is defined by an equilibrium line altitude (ELA) and a mass balance gradient with altitude (in [mm m-1]). Above the ELA, we add ice (from snow) and below the line we remove ice (melting it). We will learn more about this in the accumulation and ablation notebook.

The mass balance model mb_model gives us the mass balance for any altitude we want. We will plot it below.

Glacier initialisation

|\colorbox{yellow}{We can now take our bed and the mass balance and create a glacier which we can then perform experiments on.}|

Similarly to the bed, the Glacier object is storing some information that we can recover just by calling it:
\end{lstlisting}
\caption{Example text snippet from the climate change dataset. The phrase highlighted in line 7 describes the instructions highlighted in listing \ref{listing:source-climate}.}
\label{listing:text-climate}
\end{listing*}

\section{Qualitative Analysis of the Generated Descriptions}\label{appendix:analysis}
We conducted a manual analysis of the quality of the code descriptions generated using the prompt from listing \ref{listing:prompt}. The evaluation was performed over a random sample of 20 code snippets, which are either functions, class definitions of complete source files, alongside their natural language explanation. Each explanation consist of multiple phrases, where each phrase is aligned to a specific line or line span of the corresponding source code snippet. The annotator inspected the pair and determined whether the automatically generated description was good, if the description narrates the semantics of the code accurately; acceptable, if it is technically correct but too close to the code's level of abstraction; or bad when the description is incorrect or inaccurate. Out of the 20 pairs, 85\% (17 pairs) were deemed of good quality, 15\% (3 pairs) were deemed of acceptable quality and none were considered bad.
Listings \ref{listing:analysis1} through \ref{listing:analysis5} show a few examples of the annotated pairs. The complete set of annotations will be resealed as an artifact along the code and dataset upon publication.

        \begin{listing*}[!ht]
\begin{lstlisting}[escapeinside=||,
breaklines=true,
frame=lines,
framesep=2mm,
basicstyle=\footnotesize\ttfamily]
We begin by importing the necessary modules for our task [1-2]. The 'setuptools' module provides us with the 'setup' function, which is a utility for packaging Python projects [1]. The 'Cython.Build' module provides us with the 'cythonize' function, which is used to convert Python scripts into C code [2].

Next, we use the 'setup' function to package our Python project [4-8]. The 'ext_modules' argument is used to specify the Python scripts that we want to convert into C code [5-7]. We pass a list of three scripts to the 'cythonize' function: 'listdict.pyx', 'simulations.py', and 'tracer.py' [5-7]. These scripts are located in the 'pyscripts' directory [5-7]. 

Finally, we specify the language level of our Python scripts as '3' [7]. This means that our scripts are written in Python 3, and the 'cythonize' function should treat them as such when converting them into C code [7].
     \end{lstlisting}
\begin{lstlisting}[escapeinside=||,
breaklines=true,
frame=lines,
framesep=2mm,
basicstyle=\footnotesize\ttfamily,
numbers=left,numberstyle=\tiny\color{gray},
language=Python]
from setuptools import setup
from Cython.Build import cythonize

setup(
    ext_modules=cythonize(["pyscripts/listdict.pyx",
                           "pyscripts/simulations.py",
                           "pyscripts/tracer.py"],  language_level="3")
)\end{lstlisting}
\caption{Synthetic text-code training data pair annotated as {\textbf{Good}}.}
\label{listing:analysis1}
\end{listing*}

        \begin{listing*}[!ht]
\begin{lstlisting}[escapeinside=||,
breaklines=true,
frame=lines,
framesep=2mm,
basicstyle=\footnotesize\ttfamily]
The code defines a function that performs a two-sample Kolmogorov-Smirnov test, which is a nonparametric test used to compare two samples to determine if they are drawn from the same distribution [1]. 

First, the function converts the two input data sets into masked arrays, which are arrays that may have missing or invalid entries [2]. The function then calculates the number of elements in each data set [3]. The product of the number of elements in the two data sets divided by the sum of the number of elements in the two data sets is computed [4]. 

Next, the function concatenates the two data sets and sorts the combined data set using the merge sort algorithm [5-6]. A cumulative sum of the sorted data is calculated, where each element in the first data set contributes 1 divided by the number of elements in the first data set, and each element in the second data set contributes -1 divided by the number of elements in the second data set [7]. If there are duplicate values in the combined data set, the cumulative sum is adjusted accordingly [8-9].

The function then checks the type of alternative hypothesis specified by the user [11]. If the alternative hypothesis is 'two-sided', the function calculates the maximum absolute value of the cumulative sum [13] and computes the p-value using the Kolmogorov distribution [14]. If the alternative hypothesis is 'less', the function calculates the minimum value of the cumulative sum [16] and computes the p-value using an exponential distribution [17]. If the alternative hypothesis is 'greater', the function calculates the maximum value of the cumulative sum [19] and computes the p-value using an exponential distribution [20]. If the alternative hypothesis is not one of the three specified types, the function raises a ValueError [21-23].

Finally, the function returns the test statistic and the p-value [25]. The test statistic is a measure of the distance between the empirical distribution functions of the two samples, and the p-value is the probability of observing a test statistic as extreme as the one calculated, assuming the null hypothesis is true.
     \end{lstlisting}
\begin{lstlisting}[escapeinside=||,
breaklines=true,
frame=lines,
framesep=2mm,
basicstyle=\footnotesize\ttfamily,
numbers=left,numberstyle=\tiny\color{gray},
language=Python]
def ks_twosamp(data1, data2, alternative="two-sided"):
    (data1, data2) = (ma.asarray(data1), ma.asarray(data2))
    (n1, n2) = (data1.count(), data2.count())
    n = (n1*n2/float(n1+n2))
    mix = ma.concatenate((data1.compressed(), data2.compressed()))
    mixsort = mix.argsort(kind='mergesort')
    csum = np.where(mixsort < n1, 1./n1, -1./n2).cumsum()
    if len(np.unique(mix)) < (n1+n2):
        csum = csum[np.r_[np.diff(mix[mixsort]).nonzero()[0],-1]]

    alternative = str(alternative).lower()[0]
    if alternative == 't':
        d = ma.abs(csum).max()
        prob = special.kolmogorov(np.sqrt(n)*d)
    elif alternative == 'l':
        d = -csum.min()
        prob = np.exp(-2*n*d**2)
    elif alternative == 'g':
        d = csum.max()
        prob = np.exp(-2*n*d**2)
    else:
        raise ValueError("Invalid value for the alternative hypothesis: "
                         "should be in 'two-sided', 'less' or 'greater'")

    return (d, prob)\end{lstlisting}
\caption{Synthetic text-code training data pair annotated as {\textbf{Good}}.}
\label{listing:analysis2}
\end{listing*}
        
        \begin{listing*}[!ht]
\begin{lstlisting}[escapeinside=||,
breaklines=true,
frame=lines,
framesep=2mm,
basicstyle=\footnotesize\ttfamily]
The code begins by importing necessary libraries and modules, including the functools library, the sentry_sdk library for error tracking, the loguru library for logging, and modules related to a Telegram bot [1-4, 6-7]. 

A function decorator named 'service_template' is defined [10]. This decorator is designed to wrap around other functions and provide additional functionality. The decorator takes a function 'f' as an argument [10]. Inside the decorator, a new function 'inner' is defined, which takes two arguments: a token and a message [12]. 

When the 'inner' function is called, it attempts to execute the function 'f' with the provided token and message [14]. If the execution of 'f' raises a 'DingBotException', the exception is logged and captured by Sentry, with a specific message indicating a sending error in DingTalk [15-17]. If any other type of exception is raised, it is also logged and captured by Sentry [18-20]. If no exceptions are raised, a log message is generated indicating that the message has been sent [22]. The 'inner' function is then returned by the decorator [24].

A class named 'DingBotMessageService' is defined [27]. Inside this class, a static method 'send_text' is defined, which is decorated with the previously defined 'service_template' decorator [29-30]. This method takes a token and a message as arguments [31]. If either the token or the message is not provided, a ValueError is raised with a specific error message [32]. If both the token and the message are provided, the method attempts to send a text message using the 'ding_bot_client' [34].
     \end{lstlisting}
\begin{lstlisting}[escapeinside=||,
breaklines=true,
frame=lines,
framesep=2mm,
basicstyle=\footnotesize\ttfamily,
numbers=left,numberstyle=\tiny\color{gray},
language=Python]
import functools

import sentry_sdk
from loguru import logger

from telegram_bot.intergration import ding_bot_client
from telegram_bot.intergration.exceptions import DingBotException


def service_template(f):
    @functools.wraps(f)
    async def inner(token: str, msg: str):
        try:
            await f(token, msg)
        except DingBotException as e:
            logger.error(e)
            sentry_sdk.capture_exception(e, "DingTalk send error")
        except Exception as e:
            logger.error(e)
            sentry_sdk.capture_exception(e)
        else:
            logger.info(f"message send to {token}, msg: {msg}")

    return inner


class DingBotMessageService:

    @staticmethod
    @service_template
    async def send_text(token: str, msg: str):
        if not token or not msg:
            raise ValueError(f"Invalid param,token:{token},msg:{msg}")
        await ding_bot_client.send_text(token, msg)\end{lstlisting}
\caption{Synthetic text-code training data pair annotated as {\textbf{Good}}.}
\label{listing:analysis3}
\end{listing*}
        
        \begin{listing*}[!ht]
\begin{lstlisting}[escapeinside=||,
breaklines=true,
frame=lines,
framesep=2mm,
basicstyle=\footnotesize\ttfamily]
We define a function that takes three parameters: L_chan, a_chan, and b_chan [1]. Within this function, we create a name scope for better graph organization in TensorBoard [2]. We then return a tensor that stacks the results of three operations [3]. The first operation scales the L_chan by adding 1, dividing by 2, and multiplying by 100 [3]. The second and third operations scale the a_chan and b_chan by multiplying them by 110, respectively [3]. The stacking operation is performed along the fourth axis (axis=3) [3]. This function is used to deprocess the L*a*b* color space values back to their original range.
     \end{lstlisting}
\begin{lstlisting}[escapeinside=||,
breaklines=true,
frame=lines,
framesep=2mm,
basicstyle=\footnotesize\ttfamily,
numbers=left,numberstyle=\tiny\color{gray},
language=Python]
def deprocess_lab(L_chan, a_chan, b_chan):
    with tf.name_scope("deprocess_lab"):
        return tf.stack([(L_chan + 1) / 2 * 100, a_chan * 110, b_chan * 110], axis=3)\end{lstlisting}
\caption{Synthetic text-code training data pair annotated as {\textbf{Acceptable}}.}
\label{listing:analysis4}
\end{listing*}

\begin{listing*}[!ht]
\begin{lstlisting}[escapeinside=||,
breaklines=true,
frame=lines,
framesep=2mm,
basicstyle=\footnotesize\ttfamily]
We begin by defining a class named Column [1]. This class is initialized with a title and a list of values [3-5]. Additionally, two other attributes are initialized: 'prio' is set to 0 [6], and 'diff' is set to -1 [7].

The class includes a method that allows for comparison between two instances of the Column class [9-14]. This comparison is based on the 'prio', 'diff', and 'title' attributes of the instances [10-13].

The class also includes a method to retrieve a specific value from the 'values' attribute based on an index [16-17], and a method to set a specific value in the 'values' attribute based on an index and a new value [19-20].

Another method in the class is used to compute the number of unique values in a subset of the 'values' attribute [22-29]. This subset is determined by a list of indices [23]. The number of unique values is then stored in the 'diff' attribute [25]. If an error occurs during this process, the type of the subset and the first 10 elements of the subset are printed, and the error is raised again [24, 26-29].

Finally, the class includes a method to represent an instance of the Column class as a string [31-32]. This string includes the 'title', 'values', 'prio', and 'diff' attributes of the instance [32].
     \end{lstlisting}
\begin{lstlisting}[escapeinside=||,
breaklines=true,
frame=lines,
framesep=2mm,
basicstyle=\footnotesize\ttfamily,
numbers=left,numberstyle=\tiny\color{gray},
language=Python]
class Column(object):

    def __init__(self, title, values):
        self.title = title
        self.values = values
        self.prio = 0
        self.diff = -1

    def __lt__(self, other):
        return (self.prio, self.diff, self.title) < (
            other.prio,
            other.diff,
            other.title,
        )

    def value(self, i):
        return self.values[i]

    def set_value(self, i, v):
        self.values[i] = v

    def compute_differences(self, idx):
        x = [self.values[i] for i in idx]
        try:
            self.diff = len(set(x))
        except Exception:
            print(type(x))
            print(x[:10])
            raise

    def __repr__(self):
        return "Column(%s,%s,%s,%s)" % (self.title, self.values, self.prio, self.diff)
\end{lstlisting}
\caption{Synthetic text-code training data pair annotated as {\textbf{Acceptable}}.}
\label{listing:analysis5}
\end{listing*}

\end{document}